\documentclass[letterpaper]{article} 
\usepackage{aaai24}

\usepackage{times}  
\usepackage{helvet}  
\usepackage{courier}  
\usepackage[hyphens]{url}  
\usepackage{graphicx} 
\usepackage{natbib}  
\usepackage{caption} 
\usepackage{algorithm}
\usepackage{algorithmic}
\usepackage{xcolor}
\usepackage{color}
\usepackage{url}

\usepackage{multirow}
\usepackage{makecell}
\usepackage{amssymb}
\usepackage{pifont}

\usepackage{newfloat}
\usepackage{listings}
\DeclareCaptionStyle{ruled}{labelfont=normalfont,labelsep=colon,strut=off} 
\floatstyle{ruled}
\newfloat{listing}{tb}{lst}{}
\floatname{listing}{Listing}
\usepackage{hyperref} 

\title{SuperCLUE: A Comprehensive Chinese Large Language Model Benchmark}
\author{
    Liang Xu\thanks{Equal contribution}\textsuperscript{\rm 1}, Anqi Li\thanks{Equal contribution} \textsuperscript{\rm 2},
    Lei Zhu\textsuperscript{\rm 1},
    Hang Xue\textsuperscript{\rm 1},
    Changtai Zhu\textsuperscript{\rm 1},\\
    Kangkang Zhao\textsuperscript{\rm 1},
    Haonan He\textsuperscript{\rm 1},
    Xuanwei Zhang\textsuperscript{\rm 1},
    Qiyue Kang\textsuperscript{\rm 2},
    Zhenzhong Lan\textsuperscript{\rm 2}
}
\affiliations{
    \textsuperscript{\rm 1}CLUE, \textsuperscript{\rm 2}Westlake University

    clues@cluebenchmarks.com, \\
    \{lianqi, lanzhenzhong\}@westlake.edu.cn
}

\usepackage{bibentry}

\begin{document}

\maketitle

\begin{abstract}
Large language models (LLMs) have shown the potential to be integrated into human daily lives. Therefore, user preference is the most critical criterion for assessing LLMs' performance in real-world scenarios. However, existing benchmarks mainly focus on measuring models' accuracy using multi-choice questions, which limits the understanding of their capabilities in real applications. We fill this gap by proposing a comprehensive Chinese benchmark SuperCLUE, named after another popular Chinese LLM benchmark CLUE~\citep{xu-etal-2020-clue}. SuperCLUE encompasses three sub-tasks: actual users' queries and ratings derived from an LLM battle platform (CArena), open-ended questions with single and multiple-turn dialogues (OPEN), and closed-ended questions with the same stems as open-ended single-turn ones (CLOSE). Our study shows that accuracy on closed-ended questions is insufficient to reflect human preferences achieved on open-ended ones. At the same time, they can complement each other to predict actual user preferences. We also demonstrate that GPT-4 is a reliable judge to automatically evaluate human preferences on open-ended questions in a Chinese context~\footnote{Our benchmark will be released at \url{https://www.CLUEbenchmarks.com}}.
\end{abstract}

\section{Introduction}
Recently, large language models (LLMs) have exhibited remarkable capabilities in handling diverse problems in general and specialized domains~\citep{openai2023gpt4, zeng2022glm, du2022glm, touvron2023llama, vicuna2023, zhang2022opt, sun2023moss}. The great advancements of artificial intelligence have paved the way to the realization of integrating LLMs into human daily lives~\citep{kasneci2023education_survey, nov2023chatgpt_medical, sallam2023chatgpt_in_healthcare}. Therefore, understanding LLMs' performances on a diverse set of abilities in real-world scenarios is essential for identifying the strengths and weaknesses of the models.

The most standard and authoritative indicators for evaluating models' performances are the ratings from real users. To investigate the performance of English models from the user perspective, ~\citet{zheng2023llm-judge} develops a crowdsourced platform Chatbot Arena to allow users to engage in conversations with two anonymous chatbots at the same time and then rate based on personal preferences. 

\begin{figure}[htbp]
\centering
\includegraphics[scale=0.36]{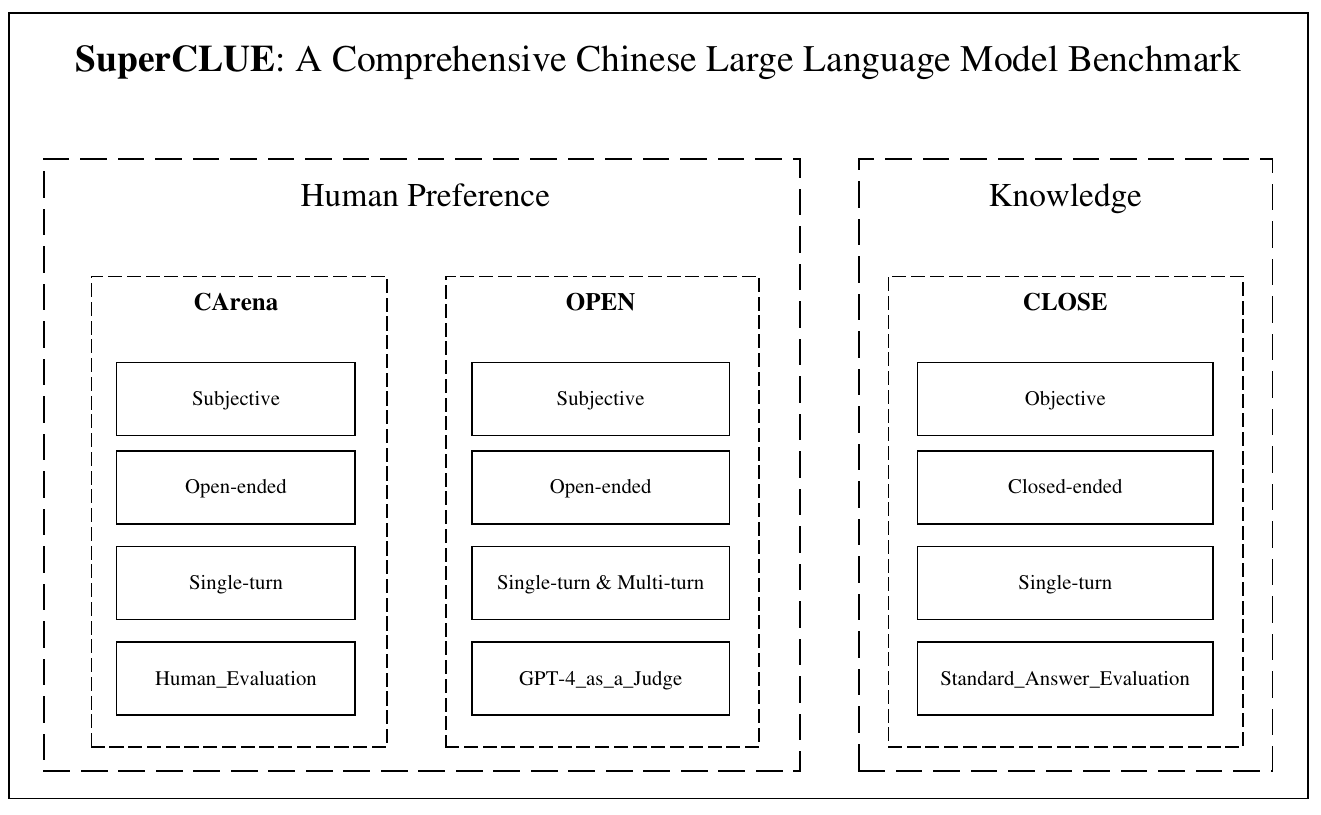}
\caption{SuperCLUE Benchmark combines three complementary evaluation methods, including open-ended questions, multiple-choice questions and side-by-side comparisons with human evaluation. }
\label{fig:superclue_benchmarks}
\end{figure}

However, such a time-consuming and cost-intensive evaluation process is almost impossible to be scaled up for each newly-developed model. 

Numerous benchmarks are proposed to assess various capabilities of the lasted LLMs efficiently. For example, MMLU~\citep{Dan2020mmlu}, Big-Bench~\citep{Srivastava2022BIGbench}, HELM~\citep{liang2022helm} and MMCU~\citep{zeng2023mmcu} are widely applied in recent research to holistically evaluate LLMs on multiple NLP tasks. Some other benchmarks like AGIEval~\citep{Zhong2023AGIEval} and C-Eval~\citep{huang2023ceval} emphasize the importance of assessing LLMs' emergent capabilities on human-level tasks, which can reflect models' real-world applicability. In addition, ~\citet{zheng2023llm-judge} proposes an English benchmark MT-bench to evaluate models' multi-turn conversational and instruction-following abilities. These benchmarks mainly consist of closed-ended multi-choice questions except MT-bench. Considering that users usually query in an open-ended manner, benchmarking with closed-ended questions may result in a limited understanding of LLMs' capabilities to interact with users in actual applications. Moreover, the correlation between capabilities reflected on all these benchmarks and perceived by real users has not been studied.

In this work, we fill the above gaps by constructing a benchmark that predicts LLMs' performances on a diverse set of abilities in real Chinese scenarios. Similar to Chatbot Arena, we design an anonymous battle platform based on the Elo rating system~\footnote{The Elo rating system calculates the relative skill levels of players in zero-sum games.}, in which users can communicate with two Chinese LLMs and rank their responses. In total, we collect 9.9k queries with ratings from real users and regard the average rate of win and tie as the gold standard of models' performance. In order to analyze Chinese users' interests, we carefully annotate part of the queries with ten capability categories, including reasoning and creation. We then construct the open-ended sub-benchmark by selecting 30 single-turn and 30 multi-turn questions for each category. And for each single-turn open-ended question, we get a corresponding closed-ended question by generating four answers using GPT-3.5 and manually checking the answers to make sure there is one and only one best choice. 

We then evaluate 11 advanced LLMs on our benchmark. To evaluate open-ended questions automatically, we use GPT-4 as the judge. Results show that GPT-4 significantly outperforms all models. Among Chinese models, MiniMax stands out and complements ChatGLM2-6B in specific categories. We also conducted extensive additional analyses on these datasets. We show that a superior large language model such as GPT-4 can be utilized as a reliable automatic rater in a Chinese context. We then analyze the correlation between GPT-4 ratings on open-ended questions and closed-ended accuracy. Based on the results, it is clear that the latter method has limitations when reflecting human preferences in open interactive scenarios. Finally, we demonstrate jointly that using close- and open-ended questions can more reliably reflect actual user preferences.

\section{Related Work}
Traditional NLP benchmarks~\citep{dolan-brockett-2005-MRPC, socher2013SST, rajpurkar2016squad, Williams2018MNLI} are mainly designed to evaluate models' performances on one specific task, such as SST-2~\citep{socher2013SST} for sentiment analysis and SQuAD~\citep{rajpurkar2016squad} for reading comprehension. To encourage the development of unified models that can handle various NLP tasks, comprehensive benchmarks to provide general-purpose evaluation~\citep{paul2019superglue, Dan2020mmlu} like GLUE~\citep{wang-etal-2018-glue} and CLUE~\citep{xu-etal-2020-clue} become prominent evaluation frameworks. Such benchmarks have significantly influenced the notable progress of generalization capabilities of language models~\citep{devlin-etal-2019-bert, brown2020gpt, zhuang-etal-2021-roberta}. Despite their broad application, most of them are not suitable for assessing recent large language models which have performed robust abilities on complex reasoning and problem-solving tasks~\citep{openai2023gpt4, Zhong2023AGIEval, huang2023ceval}. For example, \citet{goyal2023gpt-sum} show that LLMs like GPT-3 can generate more desirable summaries than "gold answers" in typical text summarization benchmarks including CNN/DailyMail.

To better understand LLMs' strengths and limitations, new benchmarks are proposed to assess broader knowledge and advanced abilities~\citep{liang2022helm, Zhong2023AGIEval, huang2023ceval, alpaca_eval}.~\citet{liang2022helm} presents a holistic evaluation of language models by taxonomizing potential application scenarios and evaluation metrics. Recent benchmarks place more emphasis on assessment in human-centric scenarios by collecting high-standard admission and qualification exams on testing human-level reasoning and real-world relevance~\citep{Zhong2023AGIEval, huang2023ceval, zeng2023mmcu, gu2023xiezhi}. For example, C-Eval~\citep{huang2023ceval} consists of exam questions that span 52 diverse disciplines ranging from humanities to science and engineering. These benchmarks mainly adopt multi-choice questions and use accuracy as the evaluation metric. Some researchers suggest that natural language generation (NLG) with multi-turn interaction should be the core evaluation approach~\citep{Dan2020mmlu, zheng2023llm-judge}.~\citet{zheng2023llm-judge} introduced MT-bench and Chatbot Arena consisting of open-ended questions that evaluate multi-turn conversational and instruction-following ability, which is the best-related dataset up to now. However, all questions in ~\citep{zheng2023llm-judge} are only open-ended and in English, and the analysis mostly focuses on investigating the agreement on ratings between superior LLMs like GPT-4 and humans. In contrast, our Chinese benchmark contains questions in both formats of open- and closed-end. And we emphasize the inadequacy of only close-ended questions, and the complementarity of the joint use of open- and close-ended ones to predict human preferences on model utilities in the real world. 

\section{SuperCLUE Benchmark}
The SuperCLUE benchmark allows developers to gather quick and accurate information about users' preferences for their models in a Chinese context before putting them into applications. We first collect a dataset CArena containing user-model interactions with user-reported ratings from a model battle platform LangYa Leaderboard~\footnote{The name LangYa Leaderboard is inspired by a Chinese TV series about heroes fighting for the top spot. You can access the platform from \url{https://www.CLUEbenchmarks.com}.}. On this platform, users can communicate with two anonymous models and then rate the models. We analyze large-scale users' queries and carefully annotate a subset of data with different model capabilities. User-reported win rates are the gold standard for measuring model performance in real-life scenarios. We then construct a smaller benchmark OPEN, which contains single- and multi-turn open-ended questions. We also design a CLOSE dataset based on the Open set, targeting to analyze whether the format of multi-choice questions is a suitable alternative to an open one in evaluating model performance. We will introduce these three datasets: CArena, OPEN, and CLOSE in detail in the following.

\subsection{CArena}
Similar to Chatbot Arena~\citep{zheng2023llm-judge} designed for English users, we developed an anonymous model competition platform LangYa Leaderboard to make Chinese LLMs (including ChatGLM-130B, ChatGLM2-6B, MOSS, Ziya, and MiniMax~\footnote{Because of the limited access to other famous Chinese LLMs such as Wenxin Yiyan, 360 brain, and SparkDesk, we cannot employ these three models in our platform.}) interact with Chinese users. On this platform, users can send queries and get replies from two anonymous models selected by the Elo rating system. After each interaction, users must determine which model's answer aligns better with their expectations. The interaction and evaluation interface is shown in Figure~\ref{fig:langyabang_user_interface} in Appendix~\ref{appendix:evaluation_interfaces}. 

We collected 9.9k votes from users from the platform since May 15th, 2023. Through analyzing user interests reflected in their queries, we find that the majority of queries can be split into ten capability groups, including semantic understanding and extraction, small talk, contextual conversation, generation and creation, knowledge and encyclopedia, code generation, logic and reasoning, calculation, role-playing, and safety. 
We then annotate queries based on these ten capability categories. 


Detailed definitions and examples of capability categories and annotation processes are shown in Appendix~\ref{appendix_capability_category}. 

\subsection{OPEN Set}
\begin{figure}[htbp]
\centering
\includegraphics[scale=0.32]{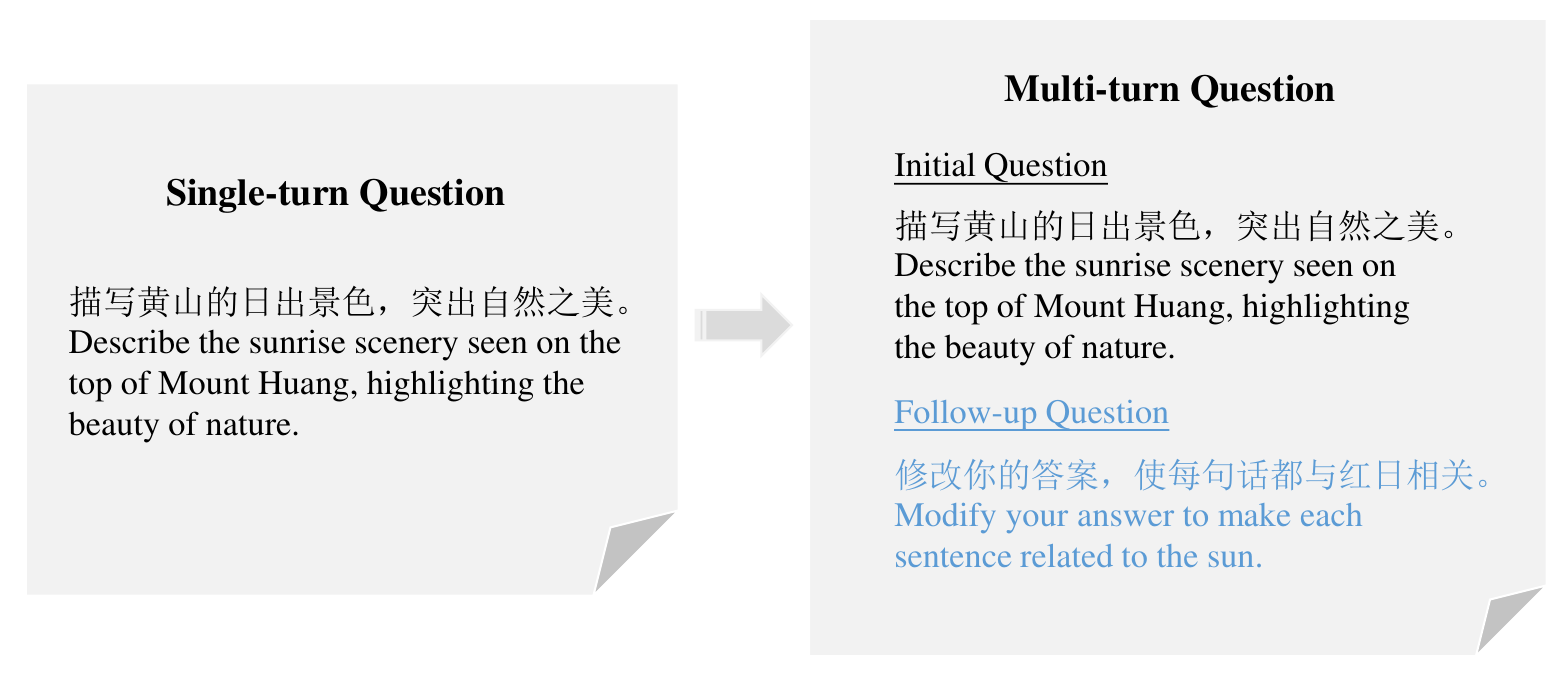}
\caption{An example question in OPEN SINGLE and its counterpart in OPEN MULTIPLE. The blue text is the follow-up question designed by humans.}
\label{fig:Open_single_and_multiple_example}
\end{figure}

Open-ended questions are questions that cannot be answered with a simple "yes" or "no", or a multi-choice, or with a specific piece of information. These questions require more thought and more than a simple one-word answer. They are designed to encourage a full, meaningful answer using the respondent's own knowledge and/or feelings.
We design the OPEN dataset following two primary principles: 1) questions in OPEN should be aligned with real users' queries in either \textbf{format} or \textbf{content}, and able to evaluate models' instruction-following abilities; 2) questions should able to evaluate models’ multi-turn conversational ability, not only single-turn; 3) abilities should be representatives and contains border categories, such as text understanding and generation, knowledge, professional and safety.
Given that users prefer interacting with models openly, we center on incorporating open-ended questions into our OPEN dataset. We then select questions based on capability categories that users concern with within the real world, with 30 single-turn questions for each category
In order to match more far-reaching application scenarios that users interact with models in multiple turns, we manually design multi-turn questions based on single-turn ones. Specifically, we regard each single-turn question as the initial question and then formulate a follow-up question (see Figure~\ref{fig:Open_single_and_multiple_example}). Finally, 600 questions are collected in total. The subset of single- or multi-turn questions in OPEN will be referred to as OPEN SINGLE or OPEN MULTIPLE in the following. single-turn and multi-turn question in OPEN will be refered to as OPEN ALL.

\subsection{CLOSE Set}
Despite its discrepancy with real queries, the format of closed-ended questions has been widely adopted by existing benchmarks due to its simple-to-evaluate properties. We attempt to quantify the limitations of closed-ended questions. 

We design a human-in-the-loop approach to transform open-ended questions in OPEN SINGLE into closed-ended ones. Specifically, we feed the stem of each open-ended question into GPT-3.5 to make it generate a four-choice question with the right answer, which is then proofread and corrected by humans. An example question in CLOSE and its counterpart in OPEN SINGLE is shown in Figure~\ref{fig:closed-ended_questions}. A more detailed process is presented in Appendix~\ref{appendix:Transformation process}. 

\begin{figure}[!htb]
\centering
\includegraphics[scale=0.22]{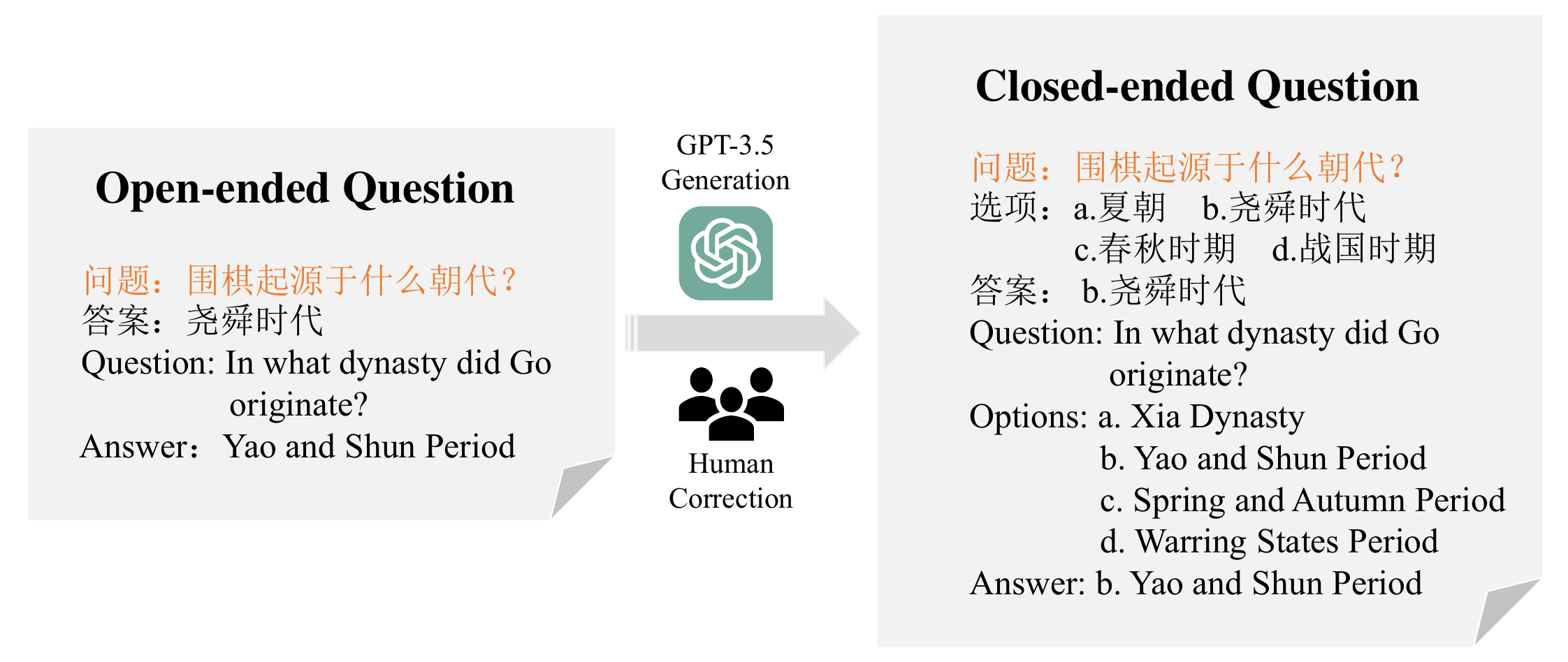}
\caption{An example question in CLOSE and its counterpart in OPEN SINGLE with the same stem (the orange text).}
\label{fig:closed-ended_questions}
\end{figure}

\section{Experiments}
In this section, we will evaluate the lasted eight Chinese-oriented LLMs on SuperCLUE, and present a comparative analysis of their performance.

\subsection{Setup}
\subsubsection{Zero-shot Evaluation}
In most real-life scenarios, models have no access to examples of specific tasks. To align with real applications, we only adopt the zero-shot setting in our evaluation. In this setting, models are evaluated on the questions without explicit training. In this setting, models are evaluated on the questions without being provided with any prior examples. 

\subsubsection{Evaluation Metrics}
For closed-ended multi-choice questions, we adopt classification accuracy as the evaluation metric. For open-ended questions and user queries collected from the model battle platform, we employ the average win rate (win and tie rate) against other models as the indicator of model performance.

\subsubsection{Evaluation Methods}
For open-ended questions, we conduct an automatic evaluation using the LLM-as-a-judge method. Specifically, we employ pairwise comparison by presenting GPT-4, serving as our default rater, with a question and answers of two models (one is the selected model, and the other is ChatGPT) and asking GPT-4 to determine which one is better or declare a tie. In the automatic evaluation process, we have mitigated certain issues of LLM judges including position bias, verbosity bias, and limited capability in grading math and reasoning questions. An example of GPT-4 judging on an open-ended question is shown in Figure~\ref{fig:contextual_dialogue_example}. Other details of automatic evaluation are presented in Appendix.

For user queries, we directly obtain user self-reported ratings from the model battle platform. The evaluation interfaces are shown in Appendix~\ref{appendix:evaluation_interfaces}.

\begin{figure*}[ht]
\centering
\includegraphics[width=0.8\textwidth]{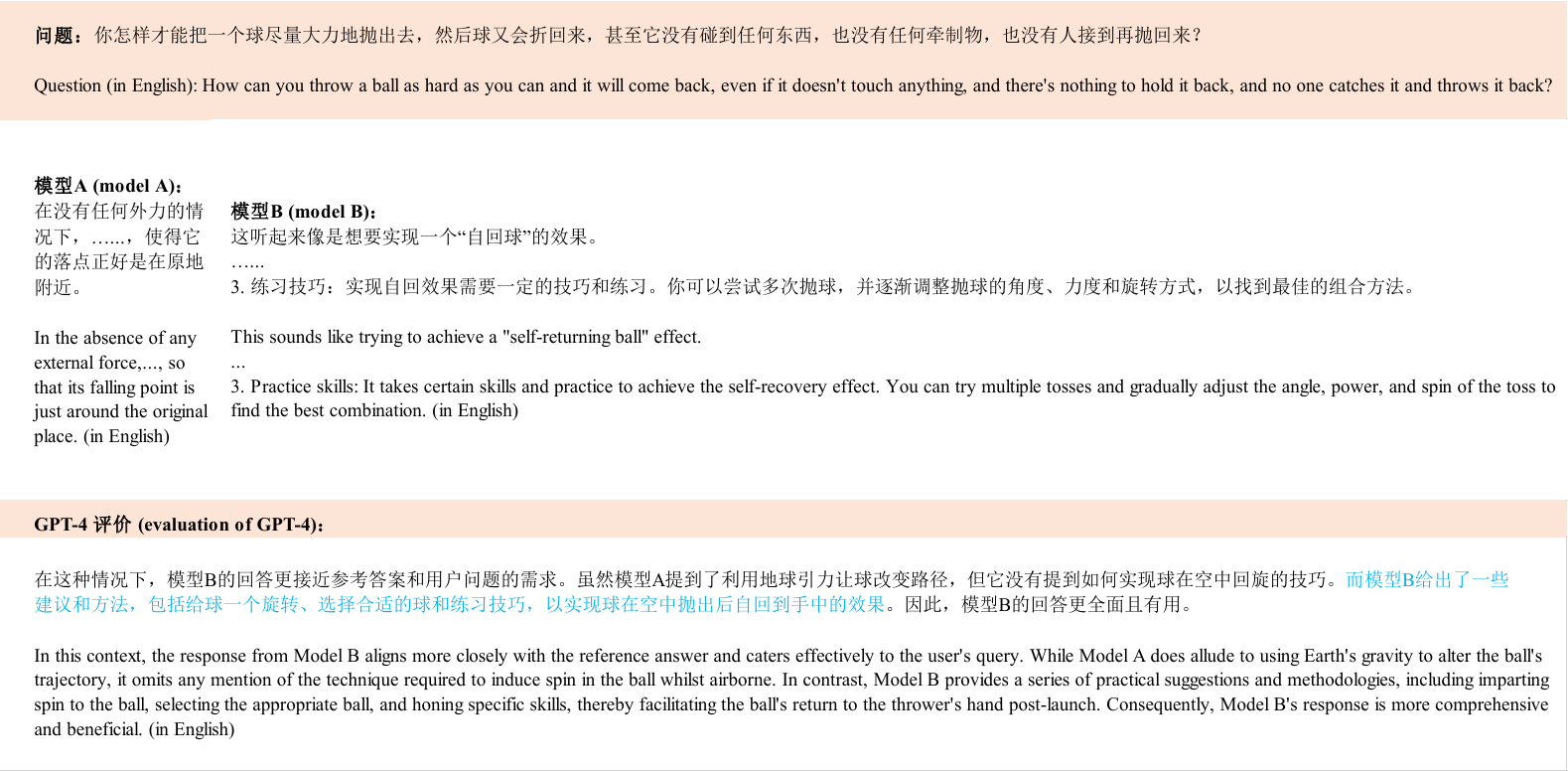}
\caption{LLM Judge example from OPEN}
\label{fig:contextual_dialogue_example}
\end{figure*}

\subsection{Models}
We focus on comprehensively evaluating the performance of three accessible LLMs that are able to process Chinese input, and eight Chinese-oriented LLMs on our benchmark, which are developed by Chinese institutions or individuals. The organizations, model sizes, and accessible approaches of the chosen models are shown in Table~\ref{tab:models_description}.

\begin{table}[]
    \centering
    \scalebox{0.8}{
    \begin{tabular}{cccc}
         \hline
         \textbf{Models} & \textbf{Developer} & \textbf{Size} & \textbf{Access} \\
         \hline
         GPT-4 & OpenAI & undisclosed & API \\
         Claude-instant-v1 & Authropic & undisclosed & API \\
         RWKV-world-7B & RWKV Foundation & 7B & Weights  \\
         ChatGLM-130B & Tsinghua & 130B & Weights \\
         ChatGLM2-6B & Tsinghua & 6B &  Weights\\
         Wenxin Yiyan (v2.0.4) & Baidu & undisclosed & API \\
         MOSS & Fudan & 16B & Weights \\
         Ziya-13B (v1.1) & IDEA & 13B & Weights \\
         360 Brain (4.0) & 360 & undisclosed & API \\
         SparkDesk (v1.5) & iFLYTEK & undisclosed & API \\
         MiniMax & MiniMax & undisclosed & API \\
         \hline
    \end{tabular}}
    \caption{LLMs chosen for evaluation. The ``size" column represents the number of parameters of each model. The ``access" column represents approaches to obtain models - through API or loading models with weights.}
    \label{tab:models_description}
\end{table}

GPT-4~\citep{openai2023gpt4} is widely known as top-performing LLM developed by OpenAI, with the training process as pretraining, instruction tuning and reinforcement learning from human feedback~\citep{ouyang2022RLHF}. Claude-instant-v1 is a light version of Claude~\footnote{\url{https://claude.ai/}}, whose performance has been demonstrated as comparable with ChatGPT~\footnote{\url{https://chat.openai.com}}. RWKV-world-7B~\footnote{\url{https://github.com/BlinkDL/ChatRWKV}} is an open-sourced RNN-based language model trained on more than 100 world languages. Among Chinese-oriented LLMs, ChatGLM-130B~\footnote{\url{https://chatglm.cn}} and ChatGLM2-6B~\footnote{\url{https://github.com/THUDM/ChatGLM2-6B}} are pre-trained using the algorithm of General Language Model~\citep{du2022glm} on bilingual (English \& Chinese) dataset and further fine-tuned on conversational data and aligned with human preferences. MOSS~\citep{sun2023moss} is considered by some the first open-source ChatGPT-like Chinese model. We evaluate the moss-moon-003-sft version in our experiment. Ziya-13B-v1.1~\citep{fengshenbang} is a large-scale pre-trained model based on LLaMA~\citep{touvron2023llama}. MiniMax~\footnote{\url{https://api.minimax.chat/}} is a newly-generated Chinese LLM based on Transformer. Wenxin Yiyan (v2.0.4)~\footnote{\url{https://yiyan.baidu.com/}}, 360 Brain (4.0)~\footnote{\url{https://ai.360.cn/}} and SparkDesk (v1.5)~\footnote{\url{https://xinghuo.xfyun.cn/}} are other Chinese LLMs whose architectures and training details are not public. 

\subsection{Results}
\begin{figure}[h]
\centering
\includegraphics[scale=0.32]{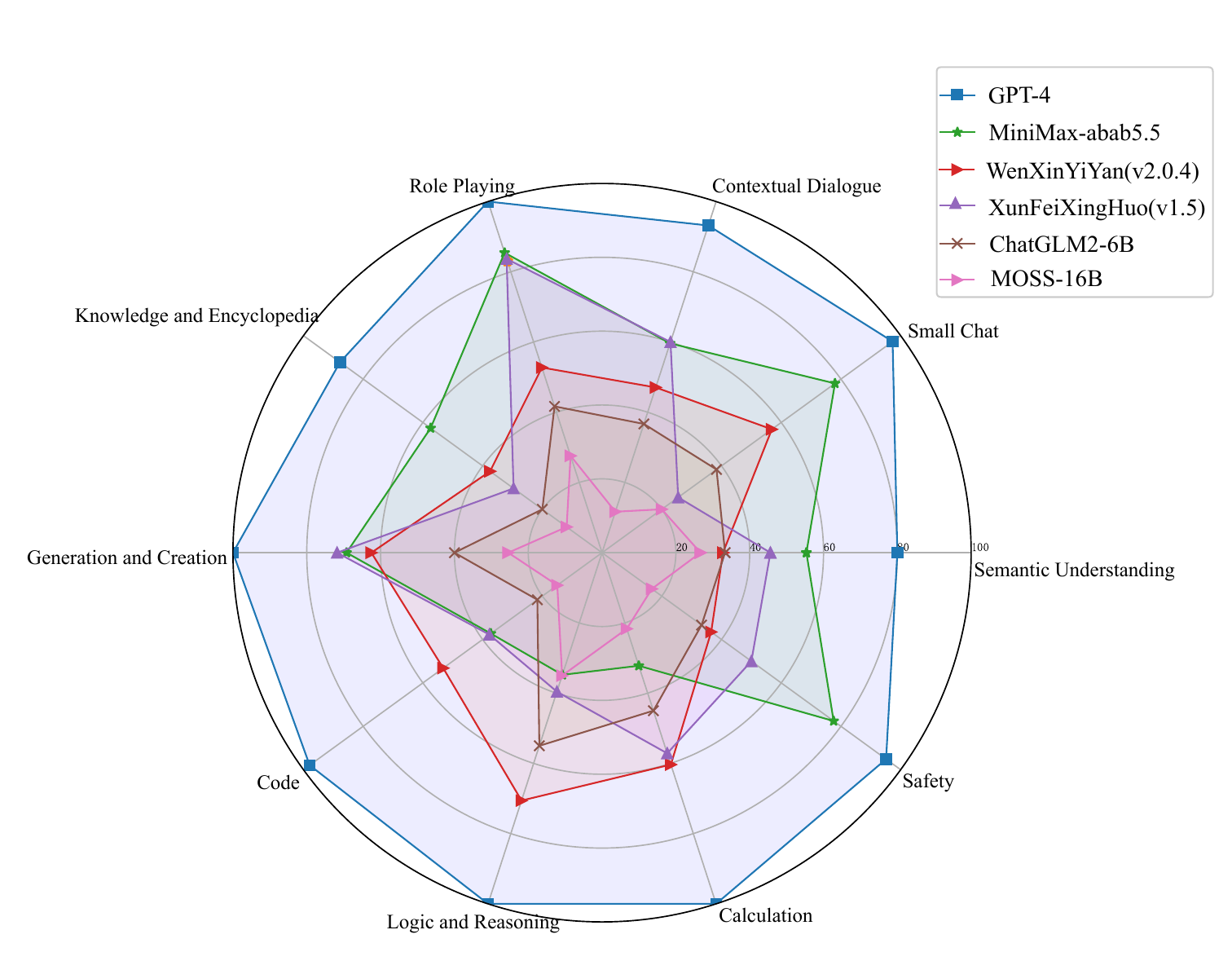}
\caption{Radar charts of capabilities distribution on OPEN of chosen LLMs.}
\label{fig:radar}
\end{figure}

The overall results of all the models and the more detailed results on each capability are provided in Table~\ref{tab:results_on_superglue} and Table~\ref{tab:results_on_superglue_capability} respectively. Radar charts that directly show model performance on each capability category are shown in Figure~\ref{fig:radar}.

\paragraph{Comparison between GPT-4 and Chinese LLMs.} GPT-4 outperforms notably  than all other models on CLOSE, OPEN SINGLE, and OPEN ALL.
MiniMax is the second-best model overall and the best-performing Chinese LLM. Compared to MiniMax, GPT-4 achieves more than ten percentage points on CLOSE and almost twice the win rate on the OPEN benchmarks. Such results indicate a large gap between Chinese-oriented models and the current top-performing models in the world. 

\paragraph{Comparison among Chinese LLMs.} MiniMax is the top model on the LangYa Leaderboard, while the second-best model ChatGLM2-6B lags behind with nearly a 1\% win rate. More specifically, Minimax beats ChatGLM2-6B in five capability categories, including Small Talk, Role Playing, Knowledge and Encyclopedia, Generation and Creation, and Logic and Reasoning. This observation suggests that complementing MiniMax with ChatGLM2 is a possible way to develop comprehensive models. We also find that all Chinese models perform similarly on the CLOSE benchmark, with seven of eight models scoring between 55\% and 60\%, while their performance on OPEN varies from 12.50\% to 41.48\%. This phenomenon suggests that close-ended multi-choice questions alone may not differentiate model capabilities.

\begin{table}[ht]
\centering
\scalebox{0.75}{
\begin{tabular}{cccccc}
\hline
\textbf{Models}   & \textbf{CLOSE} & \textbf{\makecell[c]{OPEN\\ SINGLE}} & \textbf{\makecell[c]{OPEN \\ MULTI}} & \textbf{\makecell[c]{OPEN\\ ALL}} & \textbf{CArena} \\\hline
GPT-4             & \textbf{70.67}          & \textbf{94.52}                                                 & \textbf{94.87}                                                & \textbf{94.64}                                              & -      \\
Claude-instant-v1 & 64.33          & 69.62                                                          & 69.36                                                         & 69.51                                                       & 86.00           \\
MiniMax           & 60.67          & 65.32                                                          & 47.34                                                         & 57.94                                                       & \textbf{86.69} \\
Wenxin Yiyan      & 56.67          & 57.09                                                          & 41.70                                                         & 50.48                                                       & -               \\
SparkDesk         & 55.67          & 59.52                                                          & 32.64                                                         & 48.87                                                       & -               \\
ChatGLM-130B      & 57.24          & 51.19                                                          & 30.14                                                         & 42.46                                                       & 82.76           \\
ChatGLM2-6B       & 57.67          & 42.33                                                          & 30.67                                                         & 36.50                                                       & 85.63           \\
360 Brain         & 56.38          & 28.91                                                          & 18.42                                                         & 23.93                                                       & -               \\
Ziya-13B          & 56.67          & 29.05                                                          & 14.12                                                         & 22.04                                                       & 72.48           \\
MOSS              & 41.00          & 27.00                                                          & 15.20                                                         & 21.14                                                       & 69.01           \\
RWKV-world-7B     & 19.67          & 17.17                                                          & 6.64                                                          & 12.45                                                       & 80.15           \\
 \hline
\end{tabular}}
\caption{The overall results of models' zero-shot accuracy on CLOSE, win\&tie rate on OPEN SINGLE, OPEN MULTI, OPEN ALL and CArena. The best-performing results in each column are bolded.}
\label{tab:results_on_superglue}
\end{table}

\begin{table*}[]
\centering
\scalebox{0.6}{
\begin{tabular}{cccccccccccc}
\hline
\textbf{Capability}                         & \textbf{GPT-4}  & \textbf{Claude-instant-v1} & \textbf{RWKV-world-7B} & \textbf{ChatGLM} & \textbf{ChatGLM2} & \textbf{Wenxin Yiyan} & \textbf{Moss} & \textbf{Ziya} & \textbf{360} & \textbf{SparkDesk} & \textbf{MiniMax} \\ 
\hline
\multirow{5}{*}{Semantic Understanding}     & \textbf{0.8333} & 0.6333                     & 0.1667                 & 0.4333           & 0.7667            & 0.7667                & 0.4000        & 0.6000        & 0.6667       & 0.6667             & 0.7000           \\
                                            & \textbf{0.7586} & 0.6552                     & 0.1333                 & 0.5172           & 0.4333            & 0.4667                & 0.4667        & 0.2000        & 0.4000       & 0.5333             & 0.6333  \\
                                            & \textbf{0.8462} & 0.6296                     & 0.0588                 & 0.3704           & 0.2333            & 0.1786                & 0.0667        & 0.0667        & 0.1034       & 0.3704             & 0.4615           \\
                                            & \textbf{0.8000} & 0.6429                     & 0.1064                 & 0.4464           & 0.3333            & 0.3276                & 0.2667        & 0.1333        & 0.2542       & 0.4561             & 0.5536           \\
                                            & -               & 0.8780                     & 0.2353                 & 0.7333           & \textbf{1.0000}   & -                     & 0.5000        & 0.6379        & -            & -                  & 0.7966           \\\hline
\multirow{5}{*}{Small Talk}                 & 0.6667          & \textbf{0.7667}            & 0.1000                 & 0.5667           & 0.5000            & 0.6333                & 0.4333        & 0.6000        & 0.6000       & 0.6333             & 0.6333           \\
                                            & \textbf{1.0000} & 0.9286                     & 0.1333                 & 0.6897           & 0.4667            & 0.6667                & 0.2000        & 0.1667        & 0.1724       & 0.3000             & 0.8000           \\
                                            & 0.8571          & \textbf{0.9167}            & 0.0345                 & 0.3000           & 0.3000            & 0.4286                & 0.2000        & 0.0000        & 0.1667       & 0.1905             & 0.7500           \\
                                            & \textbf{0.9730} & 0.9231                     & 0.0847                 & 0.5306           & 0.3833            & 0.5686                & 0.2000        & 0.0847        & 0.1695       & 0.2549             & 0.7800           \\
                                            & -               & 0.8578                     & 0.4757                 & 0.8412           & 0.9041            & -                     & 0.7067        & 0.7460        & -            & -                  & \textbf{0.9045}  \\\hline
\multirow{5}{*}{Contextual Dialogue}        & 0.9000          & \textbf{0.9667}            & 0.1333                 & 0.8330           & 0.7333            & 0.6000                & 0.5667        & 0.7667        & 0.7333       & 0.7333             & 0.8333           \\
                                            & \textbf{0.9333} & 0.7586                     & 0.1333                 & 0.6000           & 0.3333            & 0.5333                & 0.1667        & 0.3448        & 0.2500       & 0.6000             & 0.6667           \\
                                            & \textbf{0.9286} & 0.6000                     & 0.1250                 & 0.3500           & 0.4000            & 0.3810                & 0.0667        & 0.1429        & 0.2222       & 0.0000             & 0.5000           \\
                                            & \textbf{0.9318} & 0.6852                     & 0.1296                 & 0.5000           & 0.3667            & 0.4706                & 0.1167        & 0.2456        & 0.2364       & 0.6000             & 0.5962           \\
                                            & -               & 0.8182                     & 0.4444                 & 0.6364           & \textbf{1.0000}   & -                     & 0.9091        & 0.6842        & -            & -                  & 0.8500           \\\hline
\multirow{5}{*}{Role Playing}               & 0.7000          & \textbf{0.9000}            & 0.3000                 & 0.7241           & 0.6667            & 0.5667                & 0.4667        & 0.7000        & 0.6667       & 0.6333             & 0.8333           \\
                                            & \textbf{1.0000} & 0.8621                     & 0.1111                 & 0.7000           & 0.4333            & 0.7000                & 0.2333        & 0.2069        & 0.1667       & 0.8966             & 0.8966           \\
                                            & \textbf{1.0000} & 0.7917                     & 0.0357                 & 0.2727           & 0.4000            & 0.3200                & 0.3214        & 0.1111        & 0.1111       & 0.7500             & 0.7895           \\
                                            & \textbf{1.0000} & 0.8302                     & 0.0727                 & 0.5192           & 0.4167            & 0.5273                & 0.2759        & 0.1607        & 0.1404       & 0.8367             & 0.8542           \\
                                            & -               & 0.8462                     & 0.3750                 & 0.7037           & 0.0000            & -                     & 0.7826        & 0.6571        & -            & -                  & \textbf{0.9189}  \\\hline
\multirow{5}{*}{Knowledge and Encyclopedia} & 0.7000          & \textbf{0.7667}            & 0.1667                 & 0.5667           & 0.6667            & 0.6667                & 0.3333        & 0.5000        & 0.6000       & 0.5667             & 0.6333           \\
                                            & \textbf{0.8333} & 0.4667                     & 0.2000                 & 0.5000           & 0.2333            & 0.3793                & 0.1667        & 0.3103        & 0.1724       & 0.4000             & 0.6333           \\
                                            & \textbf{0.9474} & 0.5769                     & 0.0345                 & 0.2692           & 0.1667            & 0.3704                & 0.0690        & 0.1724        & 0.0333       & 0.1667             & 0.5000           \\
                                            & \textbf{0.8776} & 0.5179                     & 0.1186                 & 0.3929           & 0.2000            & 0.3750                & 0.1186        & 0.2414        & 0.1017       & 0.2963             & 0.5741           \\
                                            & -               & 0.8250                     & 0.4359                 & 0.8298           & 0.8155            & -                     & 0.6939        & 0.7508        & -            & -                  & \textbf{0.8863}  \\\hline
\multirow{5}{*}{Generation and Creation}    & \textbf{0.8000} & 0.5667                     & 0.2000                 & 0.6333           & 0.6667            & 0.6667                & 0.5000        & 0.7000        & 0.6333       & 0.6667             & \textbf{0.8000}  \\
                                            & \textbf{1.0000} & 0.5000                     & 0.1667                 & 0.5714           & 0.5333            & 0.6429                & 0.3667        & 0.2333        & 0.5172       & 0.7931             & 0.7333           \\
                                            & \textbf{1.0000} & 0.5294                     & 0.0000                 & 0.4167           & 0.2667            & 0.5833                & 0.1379        & 0.1429        & 0.2632       & 0.5000             & 0.5556           \\
                                            & \textbf{1.0000} & 0.5106                     & 0.1020                 & 0.5250           & 0.4000            & 0.6250                & 0.2542        & 0.1961        & 0.4167       & 0.7179             & 0.6923           \\
                                            & -      & 0.8375                     & 0.4958                 & 0.8465           & 0.8500            & -                     & 0.6532        & 0.6730        & -            & -                  & \textbf{0.8691}  \\\hline
\multirow{5}{*}{Code Generation}            & \textbf{0.6000} & 0.2000                     & 0.2667                 & 0.5172           & 0.4000            & 0.3333                & 0.3333        & 0.2333        & 0.5667       & 0.4667             & 0.3333           \\
                                            & \textbf{0.9667} & 0.4828                     & 0.3667                 & 0.2000           & 0.2333            & 0.5333                & 0.2333        & 0.3333        & 0.4000       & 0.4667             & 0.5333           \\
                                            & \textbf{1.0000} & 0.6190                     & 0.0909                 & 0.0909           & 0.2000            & 0.5294                & 0.0667        & 0.1429        & 0.2174       & 0.2609             & 0.1429           \\
                                            & \textbf{0.9792} & 0.5400                     & 0.2500                 & 0.1707           & 0.2167            & 0.5319                & 0.1500        & 0.2549        & 0.3208       & 0.3774             & 0.3725           \\
                                            & -               & \textbf{0.9756}            & 0.3725                 & 0.8302           & 0.7778            & -                     & 0.6585        & 0.6053        & -            & -                  & 0.6563           \\ \hline
\multirow{5}{*}{Logic and Reasoning}        & 0.4000          & \textbf{0.4667}            & 0.2333                 & 0.3000           & 0.3667            & 0.3667                & 0.3000        & 0.4000        & 0.3000       & 0.1667             & 0.2667           \\
                                            & \textbf{1.0000} & 0.6207                     & 0.2000                 & 0.3667           & 0.6000            & 0.7000                & 0.3667        & 0.2667        & 0.4333       & 0.4828             & 0.3667           \\
                                            & \textbf{1.0000} & 0.5556                     & 0.1500                 & 0.3810           & 0.5000            & 0.7143                & 0.3333        & 0.3000        & 0.4348       & 0.2632             & 0.3125           \\
                                            & \textbf{1.0000} & 0.5957                     & 0.1800                 & 0.3725           & 0.5500            & 0.7059                & 0.3500        & 0.2800        & 0.4340       & 0.3958             & 0.3478           \\
                                            & -               & 0.8333                     & 0.2500                 & 0.7500           & 0.6667            & -                     & 0.7222        & 0.7600        & -            & -                  & \textbf{0.9545}  \\\hline
\multirow{5}{*}{Calculation}                & \textbf{0.5333} & 0.3667                     & 0.1667                 & 0.3667           & 0.2667            & 0.3333                & 0.3333        & 0.4333        & 0.2000       & 0.2333             & 0.2333           \\
                                            & \textbf{1.0000} & 0.8000                     & 0.1333                 & 0.5333           & 0.5333            & 0.6667                & 0.2667        & 0.4333        & 0.2333       & 0.8214             & 0.3448           \\
                                            & \textbf{1.0000} & 0.8000                     & 0.1111                 & 0.3103           & 0.3667            & 0.5357                & 0.1667        & 0.3448        & 0.3667       & 0.3214             & 0.3000           \\
                                            & \textbf{1.0000} & 0.8000                     & 0.1228                 & 0.4237           & 0.4500            & 0.6034                & 0.2167        & 0.3898        & 0.3000       & 0.5714             & 0.3220           \\
                                            & -               & 0.8889                     & 0.6522                 & 0.8696           & \textbf{0.9167}   & -                     & 0.5000        & 0.7879        & -            & -                  & 0.6800           \\\hline
\multirow{5}{*}{Safety}                     & \textbf{0.9333} & 0.8000                     & 0.2333                 & 0.7931           & 0.7333            & 0.7333                & 0.4333        & 0.7333        & 0.6786       & 0.8000             & 0.8000           \\
                                            & \textbf{0.9667} & 0.9000                     & 0.1333                 & 0.4483           & 0.4333            & 0.4138                & 0.2333        & 0.4138        & 0.1379       & 0.6897             & 0.9310           \\
                                            & \textbf{0.9091} & 0.8261                     & 0.0385                 & 0.1905           & 0.2333            & 0.3043                & 0.1000        & 0.0357        & 0.0000       & 0.2381             & 0.5500           \\
                                            & \textbf{0.9512} & 0.8679                     & 0.0893                 & 0.3400           & 0.3333            & 0.3654                & 0.1667        & 0.2281        & 0.0702       & 0.5000             & 0.7755           \\
                                            & -               & 0.8000                     & 0.6111                 & 0.6875           & \textbf{1.0000}   & -                     & 0.7647        & 0.9333        & -            & -                  & 0.6500  \\\hline        
\end{tabular}}
\caption{Zero-shot results on each capability. Each result of each model corresponding to each capability is reported in the form of accuracy on CLOSE, average win rate on OPEN SINGLE, OPEN MULTI, OPEN ALL, and CArena from top to bottom. 
}
\label{tab:results_on_superglue_capability}
\end{table*}

\section{Additional Analysis}
\label{additional_analysis}
In this section, we demonstrate the high agreement between superior models like GPT-4 and human raters in a Chinese context. This validates the rationale for using GPT-4 as an alternative to human evaluators in our work. We then illustrate the limitations of the close-ended multi-choice format in evaluating model performance.

\subsection{High Agreement between GPT-4 and Human Evaluation on OPEN}
\begin{figure}[htbp]
\centering
\includegraphics[scale=0.4]{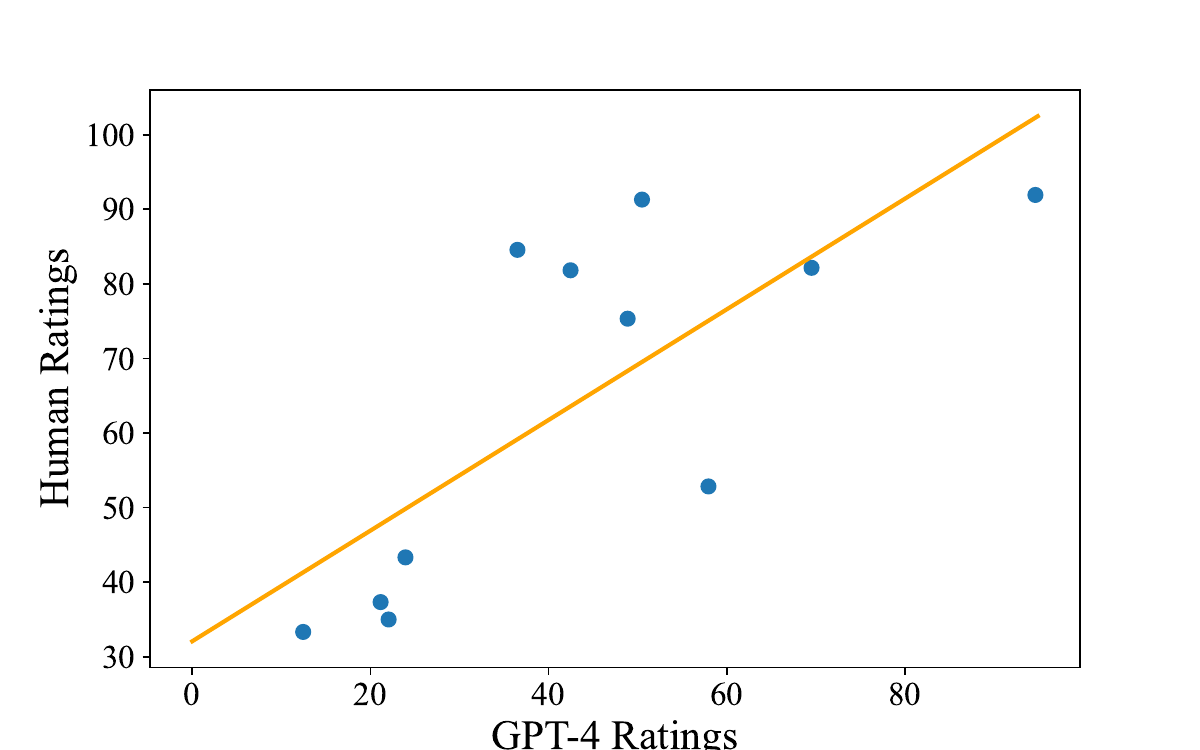}
\caption{The scatter plot of scores evaluated by GPT-4 and human, with the best-fit line.}
\label{fig:human_gpt4_agreement}
\end{figure}

We adopt GPT-4 as the default evaluator for OPEN data in our work. In order to investigate the evaluation agreement between humans and GPT-4, human raters are asked to assess a subset of model-generated answers to OPEN questions. As in the GPT-4 evaluation process, humans receive responses from one anonymous model and ChatGPT simultaneously. They must choose a better one or indicate that both are equally good. We then conduct Pearson correlation~\citep{lee1988pearson} between the average win rates evaluated by GPT-4 and human reviewers and obtain a high agreement of 80\%. This result indicates that the GPT-4 tends to closely align with those of humans. The scatter plot of the average win rate evaluated by humans and GPT-4 is shown in Figure~\ref{fig:human_gpt4_agreement}.

\subsection{Inconsistency of Chinese Model Performance on CLOSE Set and OPEN SINGLE Set}
The relationship between CLOSE Set accuracy and GPT-4 scores on OPEN SINGLE Set is analyzed by Spearman and Pearson Correlation Analysis~\citep{spearman1987spearman, lee1988pearson}. The results show that there is no significant correlation between them, with Spearman coefficient $\rho =0.5150$ with p-value $p = 0.1915$, as well as Pearson coefficient $\rho = 0.5547$ with p-value $p = 0.1536$. 

\begin{figure}
    \centering
    \scalebox{0.4}{
    \includegraphics{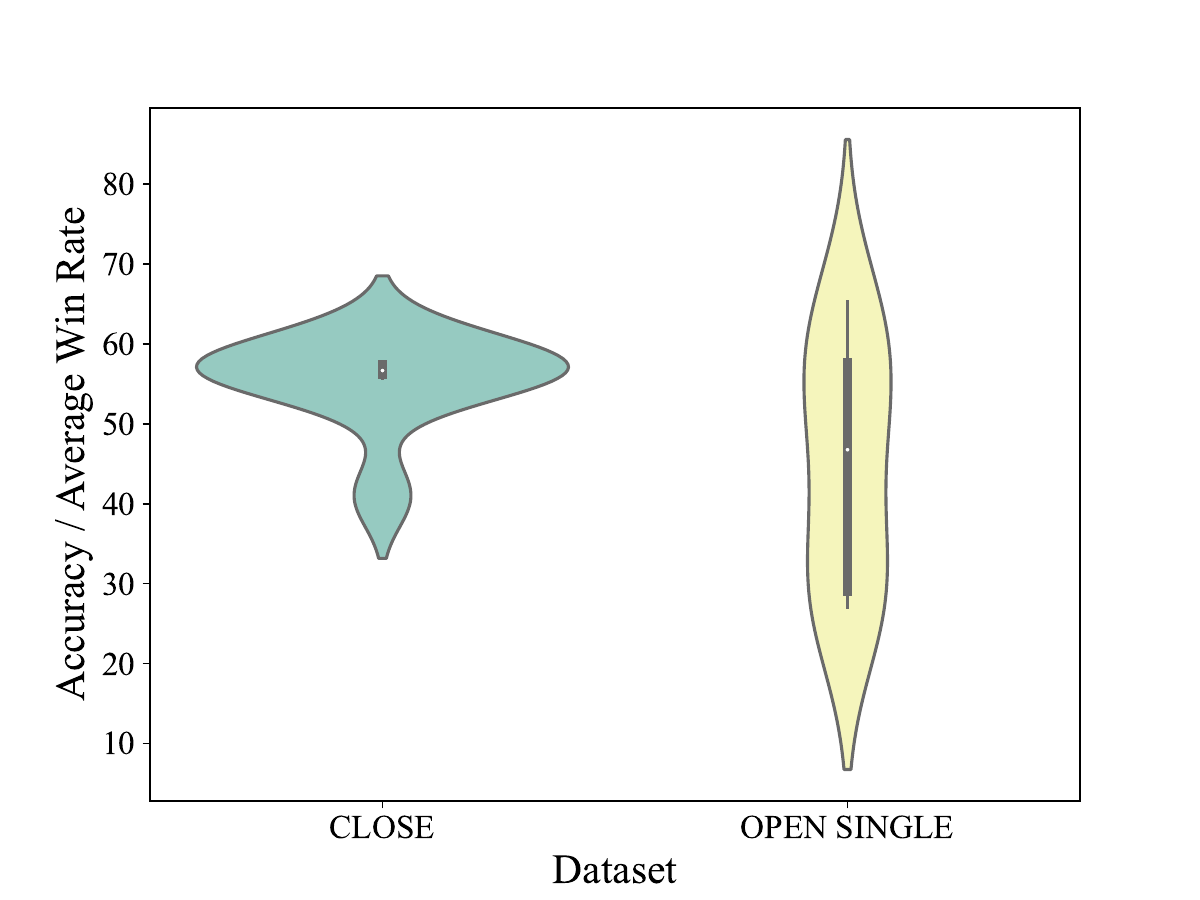}}
    \caption{The violin plot of Chinese LLMs' accuracy on CLOSE Set and average win rate on OPEN SINGLE Set.}
    \label{fig:violin_close_and_open}
\end{figure}

To further explore the relationship, we analyze the difference between the two evaluation results of Chinese LLMs in data distribution and probability density. As shown in Table~\ref{tab:results_on_superglue}, the accuracy on CLOSE is mainly concentrated between 55\% and 60\%, while the average win rate on OPEN SINGLE varies from 27.00\% to 65.32\%. Moreover, the violin plot (see Figure~\ref{fig:violin_close_and_open}) presents that the distribution of the model's performance on the OPEN SINGLE Set is much more discrete than that on the CLOSE Set, with coefficients of variation of 0.11 and 0.34 respectively. That is, closed-ended questions are not suitable as a discriminative benchmark, and cannot fully reflect model performance in open-ended interactive scenarios.


\subsection{Complementarity of the CLOSE Set and OPEN Set}
Although the above analysis demonstrates that the CLOSE Set is not as relevant as the OPEN Set for predicting model performance in the real world, we still believe that capabilities measured on the CLOSE Set could complement the OPEN Set to help understand real users' preferences better. Therefore, we combine the CLOSE, OPEN SINGLE and OPEN MULTIPLE with linear coefficients summed as 1 and make a correlation analysis with CArena. As shown in Table~\ref{tab:close_open_lyb_correlation}, accuracy on the CLOSE Set has no significant correlation with user preference in CArena, which aligns with our hypothesis and demonstration above. However, the combination with the CLOSE Set makes either OPEN SINGLE or OPEN MULTIPLE has a more significant and higher correlation with CArena than each alone. Moreover, we find another interesting phenomenon that despite users just interact with models in one turn, their preferences are more reflected from models' performance on OPEN MULTIPLE (significant correlation of $rho=0.8985$) rather than OPEN SINGLE (close to significant correlation $rho=0.8747$). These observations suggest that utilizing evaluation results on both CLOSE and OPEN MULTIPLE benefits reflecting user preference in real-life scenarios.

\begin{table}[]
    \centering
    \scalebox{0.80}{
    \begin{tabular}{lllc}
       \hline
       \textbf{\makecell[l]{CLOSE \\ Var.$|$Coef.}} & \textbf{\makecell[l]{OPEN SINGLE \\ Var.$|$Coef.}} & \textbf{\makecell[l]{OPEN MULTIPLE \\ Var.$|$Coef.}} & \textbf{Coefficient}\\ \hline
       $\checkmark$ $|$ 1 &  \ding{55} $|$ 0 & \ding{55} $|$ 0 & 0.7950 \\
        \ding{55} $|$ 0 & $\checkmark$ $|$ 1 & \ding{55} $|$ 0 & 0.8747 \\
         \ding{55} $|$ 0 & \ding{55} $|$ 0 & $\checkmark$ $|$ 1 & 0.8985*\\
        $\checkmark$ $|$ 0.5440 & $\checkmark$ $|$ 0.4560 & \ding{55} $|$ 0 & 0.9132*\\
        $\checkmark$ $|$ 0.4900 & \ding{55} $|$ 0 & $\checkmark$ $|$ 0.5100 & \textbf{0.9397}* \\
        \ding{55} $|$ 0 &  $\checkmark$ $|$ 0 & $\checkmark$ $|$ 1 & 0.8985*\\
        $\checkmark$ $|$ 0.4900 & $\checkmark$ $|$ 0 & $\checkmark$ $|$ 0.5100 & \textbf{0.9397}*\\\hline
    \end{tabular}}
    \caption{Results of Pearson correlation analysis between model performance on all possible linear combinations of the CLOSE Set, OPEN SINGLE and OPEN MULTIPLE and CArena. We report the best-fit results for each linear combination with the format of (whether the variable is selected $|$ its linear coefficient). The best linear combination with the linear coefficients of CLOSE, OPEN SINGLE and OPEN MULTIPLE are 0.49, 0, and 0.51 respectively. $*p<0.05$. The largest coefficient with a significant correlation is bolded.}
    \label{tab:close_open_lyb_correlation}
\end{table}

\section{Conclusion}
In this paper, we introduce SuperCLUE, a comprehensive benchmark for Chinese large models that includes three separate and complementary subtasks: CArena, a place where users can communicate and evaluate models; the OPEN Set, with open-ended single- and multi-turn questions; and the CLOSE Set, with closed-ended questions. We use GPT-4 for automated scoring answers to open-ended questions and demonstrate its consistency with human raters. Evaluation results show that there is still a large gap between Chinese LLMs and the top-performing models in the world. We also conducted extensive analysis to demonstrate that closed-ended questions alone cannot align with user preferences in real-world scenarios, while the joint use of closed- and open-ended questions is a better way to conduct a comprehensive evaluation. We will further expand our test set, train our evaluation model, and provide evaluation services to serve more community users.

\bibliography{custom}

\appendix
\section{Evaluation Process}
\subsection{Automatic Evaluation}
GPT-4 is required to discern which model performs better, with potential responses being 'model A is superior', 'model B is superior', or 'a tie'. Each of our topic pairs undergoes two rounds of dialogue. During the initial round, the adjudicator's evaluation is solely based on the content of this round. However, in the subsequent round, while the adjudicator is privy to the content of the initial dialogue, the evaluation primarily concentrates on the merits and demerits of the content presented in the second round.

\begin{figure*}[ht]
\centering
\includegraphics[width=1\textwidth]{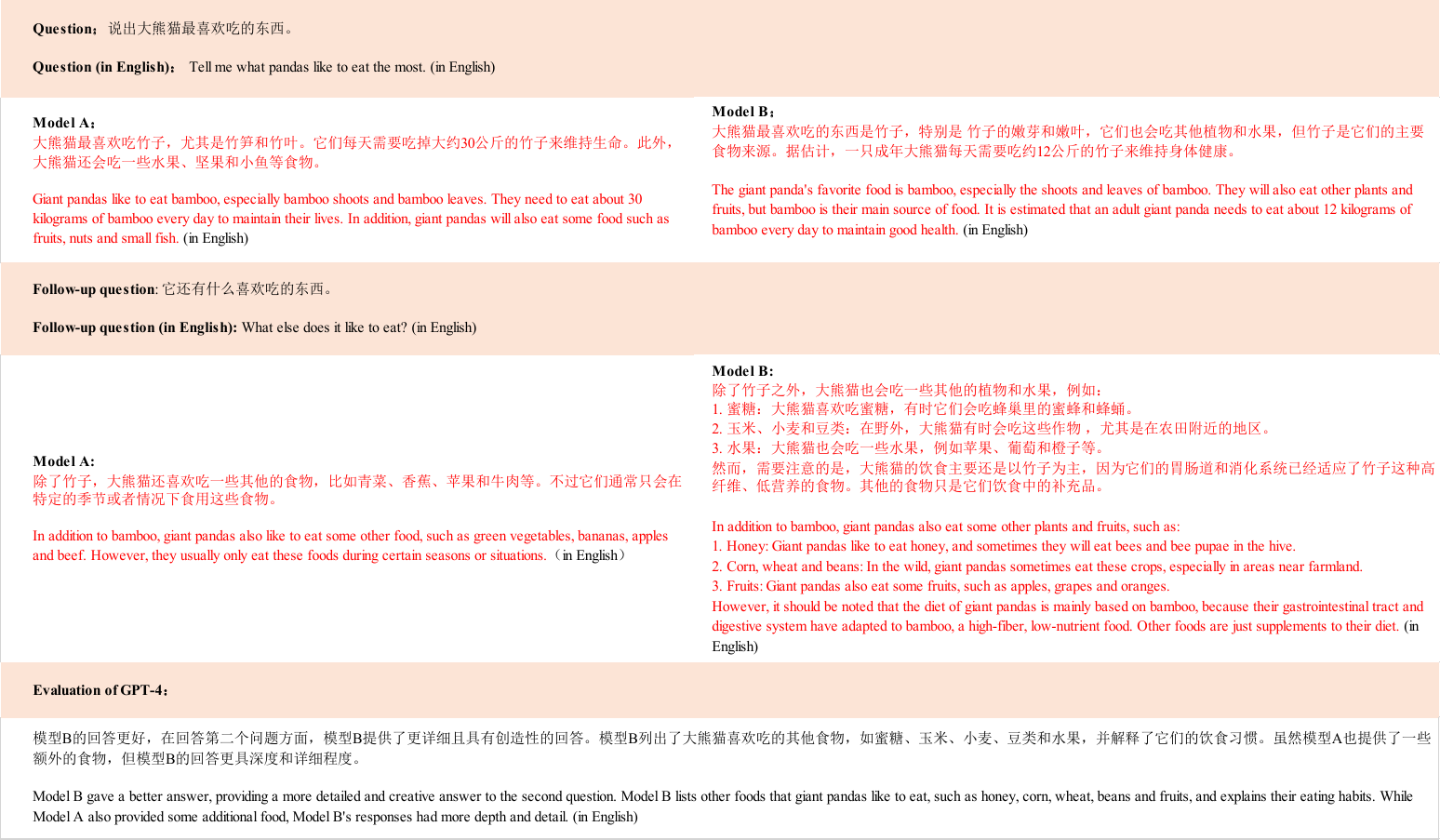}
\caption{Multi-turn dialogues between a user and two AI assistants—RWKV-world-7B (model A) and GPT-3.5-turbo (model B)——initiated by a question from the OPEN  and a follow-up instruction. A Superior model (GPT-4) is then presented with the context to determine which assistant answers better. We can see that both models can strive to do multiple-choice questions, but for open-ended follow-up questions, the quality of the answers is quite different.}
\label{fig:chat_BLM_baichuan}
\end{figure*}

\subsection{Evaluation Interfaces}
\label{appendix:evaluation_interfaces}
\begin{figure*}[ht]
\centering
\includegraphics[width=1\textwidth]{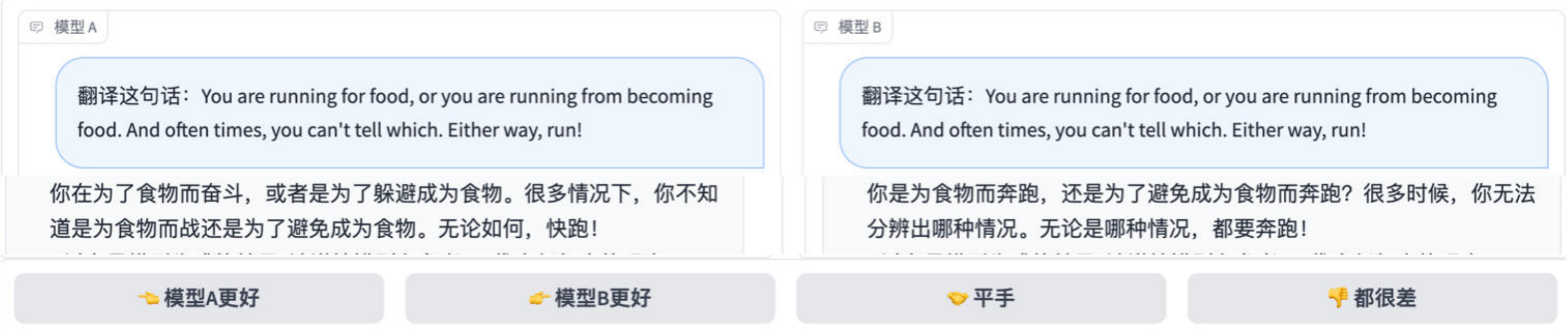}
\caption{LangYa Leaderboard user evaluation}
\label{fig:langyabang_user_interface}
\end{figure*}

The evaluation interfaces for users in the model battle platform is presented in Figure~\ref{fig:langyabang_user_interface}.

\subsection{Zero-shot Evaluation Examples}
\label{appendix:zero-shot_evaluation_examples}
\begin{figure*}[ht]
\centering
\includegraphics[width=1\textwidth]{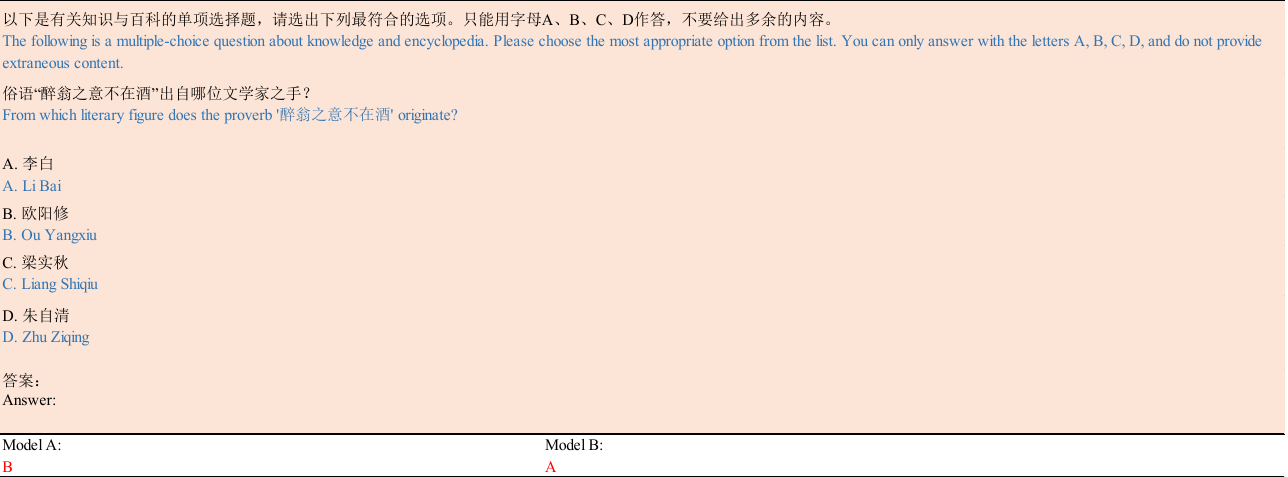}
\caption{An example of zero-shot evaluation in CLOSE. The red text is the auto-completed response from model, while the preceding text is the inputted prompt. We indicate English translation below the corresponding Chinese text.}
\label{fig:zero_shot_example_Opt}
\end{figure*}

\begin{figure*}[ht]
\centering
\includegraphics[width=1\textwidth]{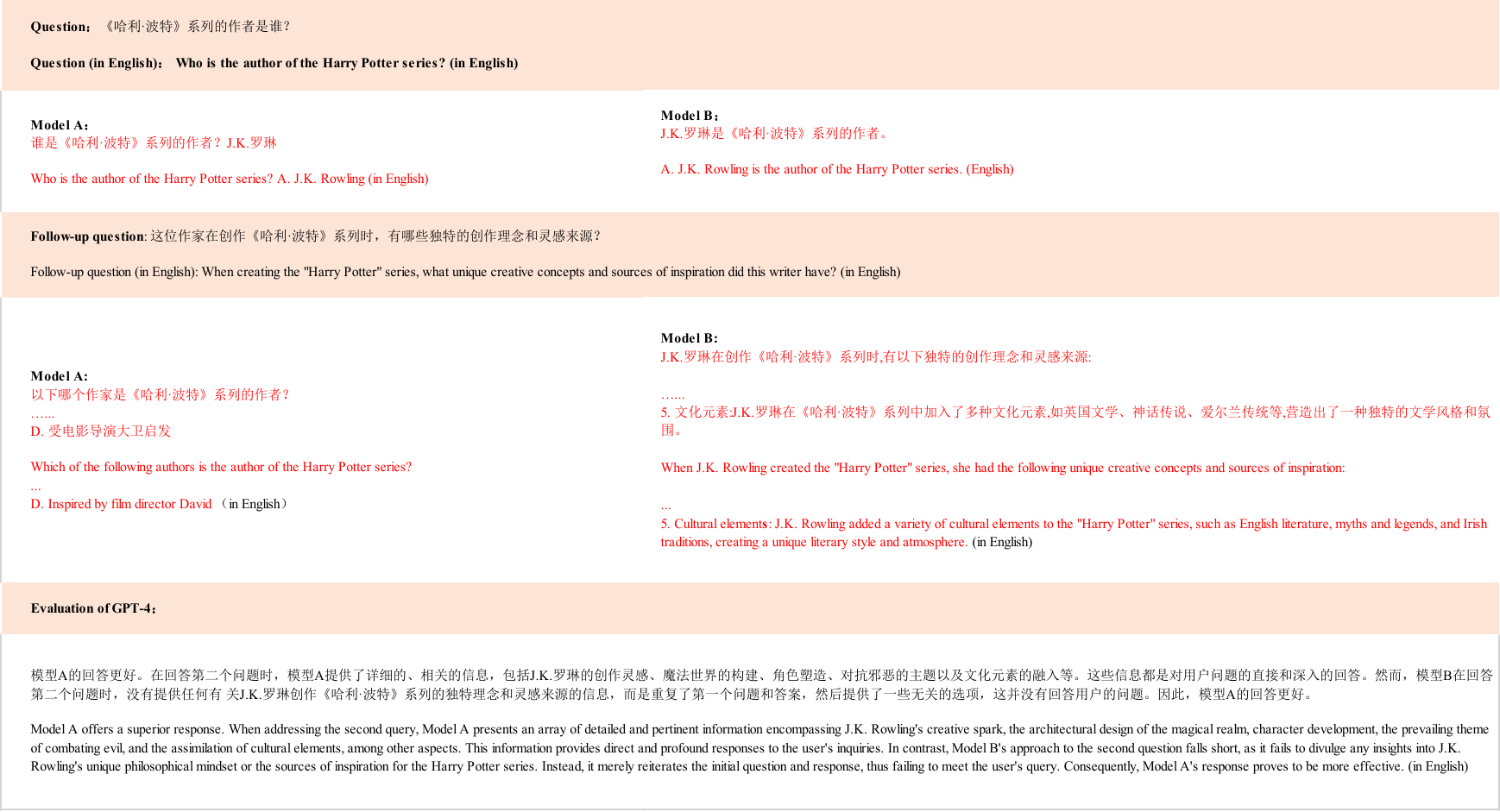}
\caption{An example of zero-shot evaluation in OPEN. The red text is the auto-completed response from model, while the preceding text is the inputted prompt. We indicate English translation below the corresponding Chinese text.}
\label{fig:zero_shot_example_Open}
\end{figure*}

The zero-shot evaluation examples of  CLOSE and OPEN are presented in Figure ~\ref{fig:zero_shot_example_Opt} and ~\ref{fig:zero_shot_example_Open}.

\subsection{Transformation Process of OPEN SINGLE to CLOSE}
\label{appendix:Transformation process}
\begin{figure*}[ht]
\centering
\includegraphics[width=1\textwidth]{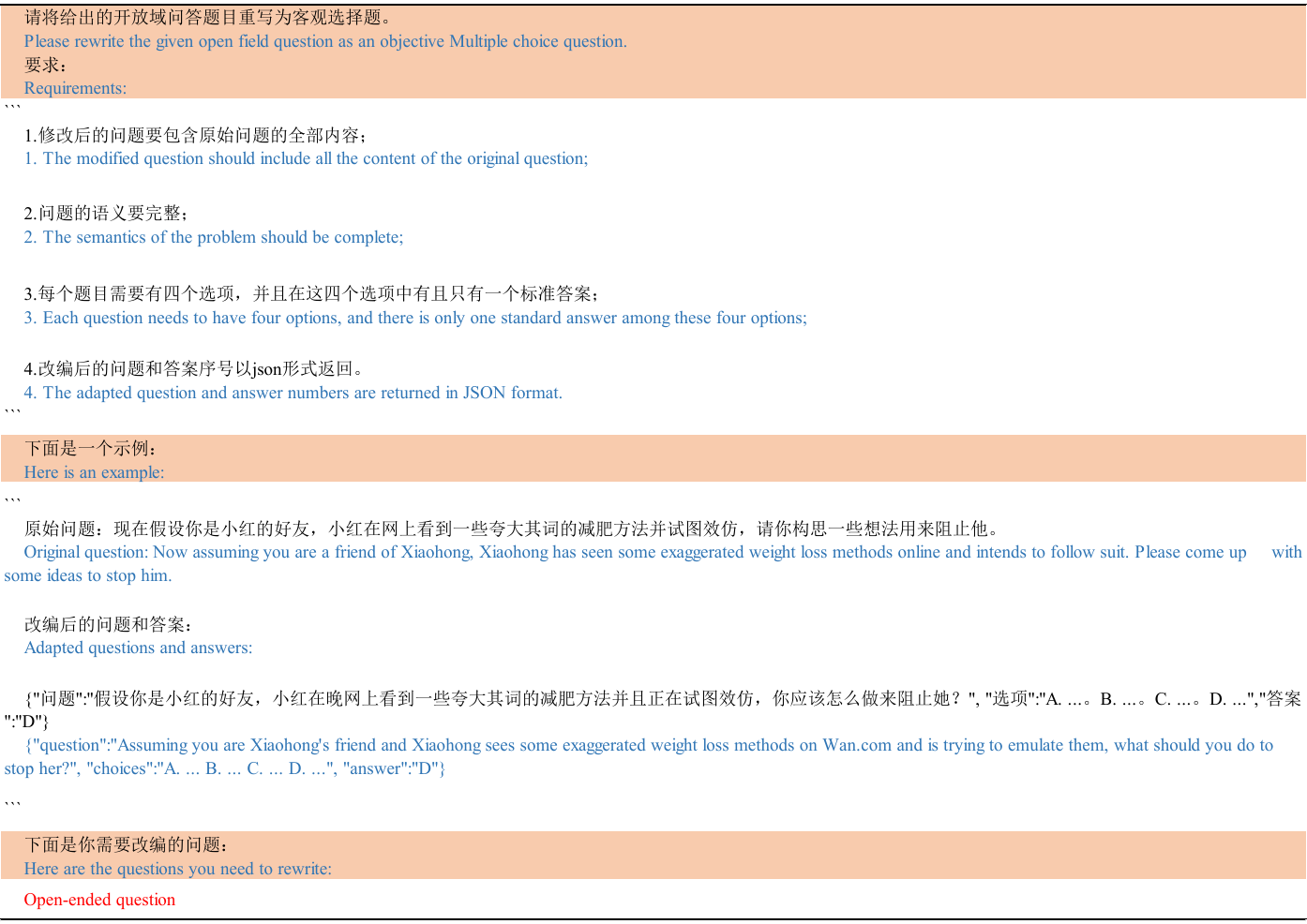}
\caption{Prompt developed for transforming open-ended questions into closed-ended ones. The red text is the open-ended question to be rewritten. We indicate English translation below the corresponding Chinese text.}
\label{fig:prompt}
\end{figure*}

Initially, by crafting a fitting prompt, we enabled GPT-3.5 to transfigure open-ended questions into closed-ended ones. The prompt we developed for transforming open-ended questions into closed-ended ones within OPEN SINGLE are visually demonstrated in Figure ~\ref{fig:prompt}. Subsequently, upon acquiring the answers generated by GPT-3.5, the reformulated questions were systematically arranged and subsequently subjected to manual evaluation to verify their compliance with the stipulated prompt requirements. Each question endured a three-stage review process, with each stage comprising a team of three human reviewers. Their collaborative task was to scrutinize and, where necessary, amend any questions that did not meet the prescribed requirements to ensure their alignment with the set guidelines.

\section{Capability Categories}
\label{appendix_capability_category}
\subsection{Detailed Definitions}
Table~\ref{tab:capability_definition_table2} presents the detailed definitions of capability categories.

\begin{table*}[ht]
\centering
\scalebox{0.8}{
\begin{tabular}{|c|c|c|}
\hline
\textbf{Primary Category} & \textbf{Secondary Category} & \textbf{Description} \\
\hline
\multirow{4}{*}{\makecell[c]{Semantic Understanding \\ and Generation}} & {\makecell[c]{Semantic Understanding \\ and Extraction}} & \makecell[l]{This refers to an ability that enables language models to understand and interpret \\the significance of entered textual information. The model must be capable of \\ identifying the meanings of phrases, sentences, and paragraphs, in addition to being \\able to extract crucial information and themes from more extensive text blocks.} \\
\cline{2-3}
 & \makecell[c]{Small Talk \\(Casual Conversation)} & \makecell[l]{This refers to the proficiency of a language model to engage in free-form, \\ non-specific-goal dialogues with users. The model needs to demonstrate the capacity \\ to generate responses that are fluent, natural, and in alignment with linguistic practices \\ and cultural nuances.} \\
\cline{2-3}
 & Contextual Conversation & \makecell[l]{This signifies a particular proficiency of the language model, necessitating the \\ understanding and retention of preceding dialogue information to maintain coherence \\ in its responses. It involves the comprehension of the overall conversation flow and \\ the surrounding context, or the generation of corresponding dialogue.} \\
\cline{2-3}
 & Generation and Creation & \makecell[l]{This denotes a distinct capability of language models, enabling them to generate novel \\ textual content, such as articles, copywriting, short stories, and poetry. It involves the \\ creative utilization of language, while also taking into consideration aspects such as \\style, context, and the target audience.} \\
\hline
\makecell[c]{Knowledge Understanding \\ and Application} & Knowledge \& Encyclopedia & \makecell[l]{This represents a distinct capability of language models, allowing them to provide \\ knowledge information akin to an encyclopedia. This encompasses understanding and \\ responding to questions about a wide array of topics, as well as providing accurate, \\ detailed, and up-to-date information.} \\
\hline
\multirow{3}{*}{Professional Abilities} & Code Generation & \makecell[l]{This refers to a unique capability of language models, enabling them to understand \\ and generate programming code. This involves the comprehension of the syntax, \\ structure, and conventions of various programming languages, as well as the method-\\ologies to solve programming problems.} \\
\cline{2-3}
 & Logic and Reasoning & \makecell[l]{This denotes a distinctive capability of language models to understand and apply \\ principles of logic for reasoning. It involves the analysis of problems, \\identification of issues,and the process of inference.} \\
\cline{2-3}
 & Calculation & \makecell[l]{This refers to a unique ability inherent in language models that enables them to perform \\ mathematical operations, such as addition, subtraction, multiplication, and division, and \\ even more complex mathematical problems. This encompasses understanding the \\ formulation of mathematical problems and the methodology to solve these problems in a \\ step-by-step manner.} \\
\hline
\multirow{2}{*}{\makecell[c]{Environmental Adaptation \\ and Safety}} & Role-playing & \makecell[l]{This denotes a distinctive capability of language models, allowing them to assume a role \\ within specific simulated environments or scenarios. This involves comprehending the \\ behavior, speaking style, and appropriate responses of a specific role under designated \\ circumstances.} \\
\cline{2-3}
 & Safety & \makecell[l]{This refers to the capability of a language model to prevent the generation of content that \\ could potentially cause distress or harm. This encompasses the identification and \\ avoidance of requests that may contain sensitive or inappropriate content, as well as \\ adherence to policies on user privacy and safety.}\\
\hline
\end{tabular}}
\caption{Detailed definitions of capability categories}
\label{tab:capability_definition_table2}
\end{table*}

\subsection{Annotation Process}
Based on the definitions and implications of the ten capability categories, we classified user queries from the LangYa Leaderboard platform. Initially, a portion of the queries were manually annotated, resulting in 300 data entries, with 30 entries dedicated to each capability category. Subsequently, these 300 annotated entries were used to train a BERT classifier. The trained classifier was then employed to classify the remaining data, assigning a specific capability category label to each query sample. Finally, a collaboration of four human evaluators reviewed and rectified the classification results produced by the BERT classifier, which led to the determination of the final capability category for each individual sample.

\end{document}